\documentclass{ecai}
\usepackage{times}
\usepackage{graphicx}
\usepackage{latexsym}
\usepackage{graphicx}
\usepackage{siunitx}
\usepackage{caption}
\usepackage{subcaption}
\newcommand{\eg}{\emph{e.g.}, }       
\newcommand{\ie}{\emph{i.e.}, }      
\newcommand{\etal}{\emph{et al.}}         
\newcommand\etc{\emph{etc.}}

\begin{document}

\title{Obstruction level detection of sewer videos using convolutional neural networks}

\author{Mario A. Guti\'errez-Mondrag\'on
\institute{Universitat Polit\`{e}cnica de Catalunya (UPC) - Barcelona Supercomputing Center (BSC), Spain, email: mario.alberto.gutierrez@upc.edu}
\and Dario Garcia-Gasulla
\institute{Barcelona Supercomputing Center (BSC), Spain,  email: dario.garcia@bsc.es}
\\\and Sergio Alvarez-Napagao
\institute{Barcelona Supercomputing Center (BSC), Spain,  email: sergio.alvarez@bsc.es}
\and Jaume Brossa-Ordo\~nez
\institute{Water Technology Center, Spain, jaume.brossa@cetaqua.com}
\and Rafael Gimenez-Esteban
\institute{Water Technology Center, Spain, rafael.gimenez@cetaqua.com}
}

\maketitle
\bibliographystyle{ecai}

\begin{abstract}

Worldwide, sewer networks are designed to transport wastewater to a centralized treatment plant to be treated and returned to the environment. This process is critical for the current society, preventing waterborne illnesses, providing safe drinking water and enhancing general sanitation. To keep a sewer network perfectly operational, sampling inspections are performed constantly to identify obstructions. Typically, a Closed-Circuit Television system is used to record the inside of pipes and report the obstruction level, which may trigger a cleaning operative. Currently, the obstruction level assessment is done manually, which is time-consuming and inconsistent. In this work, we design a methodology to train a \textit{Convolutional Neural Network} for identifying the level of obstruction in pipes, thus reducing the human effort required on such a frequent and repetitive task. We gathered a database of videos that are explored and adapted to generate useful frames to fed into the model. Our resulting classifier obtains deployment ready performances. To validate the consistency of the approach and its industrial applicability, we integrate the Layer-wise Relevance Propagation explainability technique, which enables us to further understand the behavior of the neural network for this task. In the end, the proposed system can provide higher speed, accuracy, and consistency in the process of sewer examination. Our analysis also uncovers some guidelines on how to further improve the quality of the data gathering methodology.

\end{abstract}

\section{Introduction}
In the US, there are roughly 1,200,000 kilometers of sewer lines \cite{sterling2010state}. That is more than three times the distance between the Earth and the Moon, considering only 4\% of world population. The maintenance of such vasts networks of pipes is thus a real challenge world-wide. As of now, the most common approach is to have operators executing sampling inspections, trying to find obstructions before they can cause severe failures that would require urgent and expensive actions.
The current approach is hardly scalable, as it is expensive and requires lots of human hours. Companies in charge of large wastewater networks face massive operational costs related to inspection and maintenance. The current environmental context brings added pressure to the topic, since episodes of heavy rainfall are becoming more common as a consequence of climate change \cite{donat2017addendum}. Within these episodes, obstructed wastewater networks may become the origin of sewer overflows and floods with an impact on urban environments and population.

\begin{figure}[b]
\centerline{\includegraphics[width=0.49\textwidth]{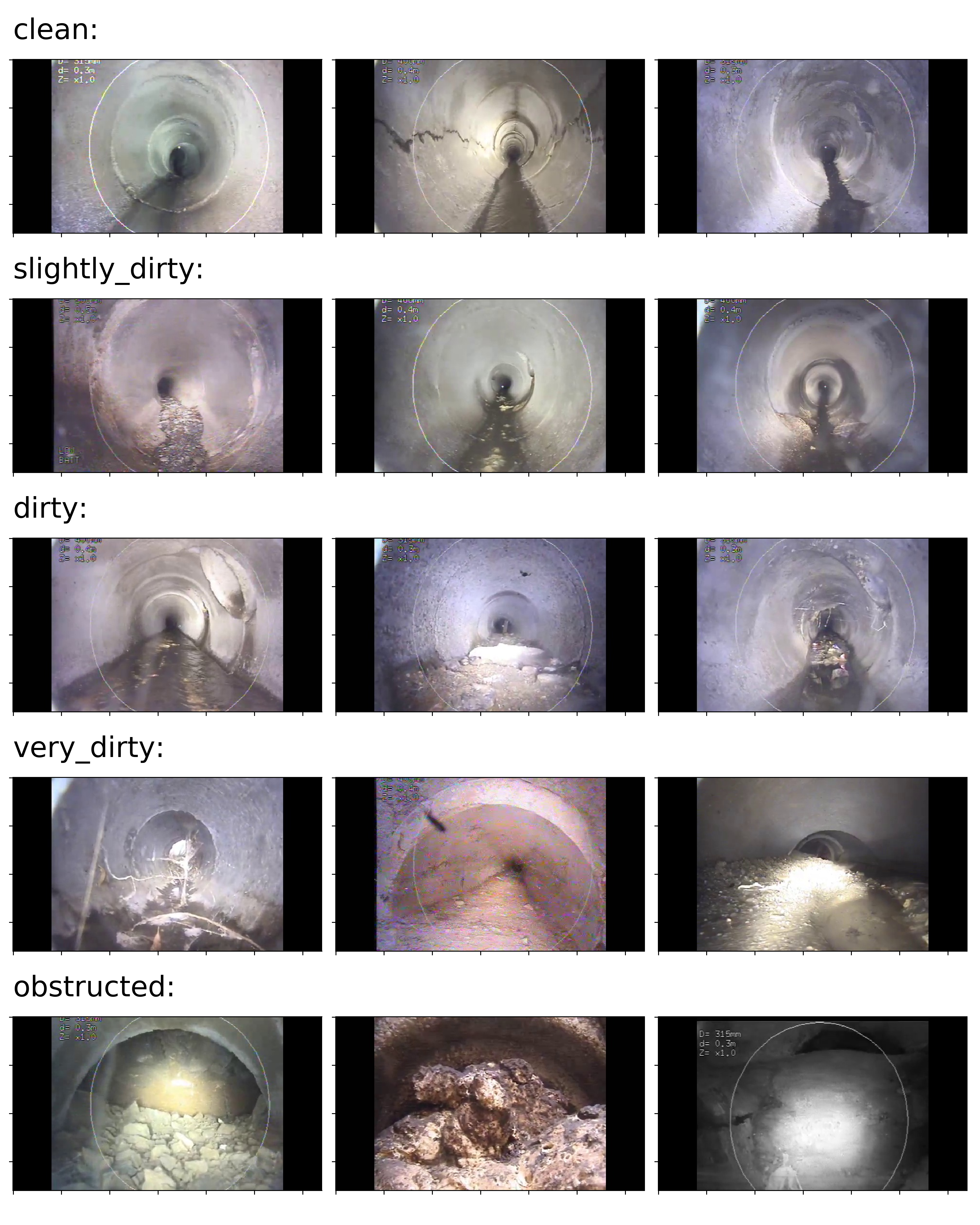}}
\caption{Sample frames from the videos database.}
\label{fig:sewer_samples}
\end{figure}

In an effort to increase the quality and efficiency of sewer maintenance, the industry is now looking into recent technological advancements in fields such as image recognition and unmanned aerial vehicles. In this paper, we tackle one of the challenges necessary for new methods to be functional: the automatic identification of obstructions in sewer pipes from image data. For this purpose we use real data from 6,590 inspection videos (samples shown in Figure \ref{fig:sewer_samples}), recorded and evaluated by operators. We post-process the videos to dissect and simplify the problem at hand. With this data we design, train and evaluate a convolutional neural network (CNN), for the task of predicting the level of obstruction of a sewer segment. The performance obtained in this work enables this technology to be directly applicable to the industrial challenge, increasing the efficiency of the procedure and enabling a more extensive maintenance. In this particular case, this is already in progress through CETaqua, industrial partner of this project, and part of the SUEZ group.

\section{Current Sewer Maintenance}\label{sec:maintenance}

Regular sewer inspections are made for the operation and maintenance of the
network. During inspections, videos of the inside of sewers are recorded by
using a camera attached to a pole. Each of these videos will be carefully
reviewed by an operator later, who must fill a report and deliver it to the
inspection site or to the central offices. The report must include the level of obstruction of the sewer, categorized into five classes: clean; slightly dirty; dirty; very dirty; and obstructed. Cleaning operations prioritize their interventions based on these reports.

Reviewing videos requires a significant amount of time from operators. This task is a major barrier for productivity because of its duration and repetitive nature; if the same operator dedicates too much time to this task, their
performance will be affected. To avoid that, in practice, many different operators end up reviewing the same videos. While this is desirable for several reasons, it entails a significant variance in the evaluation criteria. Meanwhile, a reliable and consistent evaluation is critical for the efficient planning of maintenance.

Our goal is to define and implement a system to automatically assess the obstruction on sewers from videos. This system has to provide a status on the volume of dirt or sedimentation in the pipes, to justify the cleaning needs.
The deployment of this system in production will enable a more productive use of human resources, and will provide a unified model for guiding cleaning operations.

\section{State of the Art / Related Work}


The use of computer vision techniques in civil engineering applications has grown exponentially, as visual inspections are necessary to maintain the safety and functionality of basic infrastructures. To mitigate the costs derived from the manual interpretation of images or videos, diverse studies explore the use of computer vision techniques. Methods like feature extraction, edge detection, image segmentation, and object recognition have been considered to asses the condition of bridges, asphalt pavement, tunnels, underground concrete pipes, \etc \cite{abdel2003analysis, zakeri2017image, a_rose2014supervised}. Moreover, noise reduction \cite{yamaguchi2008image}, and reconstruction and 3D visualization \cite{esquivel2009reconstruction, lattanzi20143d, huynh20163d, belles2015kinect} have also been used to improve the precision and applicability of these techniques. In the most similar scenario to the one tackled in this paper, the automatic detection of cracks in sewers has been explored through image processing and segmentation methods \cite{halfawy2013efficient, iyer2006segmentation}. 

Most of these related works are mainly focused on a single task, to detect cracks. Segmentation and classifications of pipe cracks, holes, laterals, joints and collapse surfaces are explored through mathematical morphology techniques in the work of S.K. Sinha and P. W. Fieguth \cite{sinha2006morphological}. A most recent study of L. M. Dang \etal  \cite{dang2018utilizing} uses these morphological operations and other pre-processing techniques, like edge detection and binarisation, to identify the sewer defects by recognizing text displayed on the sewer video recording. Even though computer vision techniques have provided a significant improvement in the analysis of civil infrastructure, there are still several difficulties to overcome, such as the extensive pre-processing of the data that must be carried out, a high degree of expert knowledge in the design of complex features extractors, the treatment of noisy and low-quality data, among others. In this regard, CNNs require little image pre-processing, and more importantly, the feature extraction processes is learned automatically from the data through an optimization process. The performance of CNN models has been tested in several computer vision tasks, such as object detection or image classification. For instance, in the work of Y.-J. Cha \etal \cite{cha2017deep}, an automated civil infrastructure damage system is presented, which is insensitive to the quality of the data and to camera specifications. Furthermore, CNNs use has demonstrated its efficiency in tunnel inspections \cite{montero2015past}, revealing how the deep learning approach outperforms conventional methods. 

In the case of sewer inspections, the use of neural networks has been limited to defect detection. S.S. Kumar \etal proposes a convolutional neural network to identify root intrusions, deposits, and cracks in a set of sewer videos \cite{kumar2018automated}. This database is transformed into a sequence of RGB images and fed them to the model. The training methodology is very straightforward, all images comprising a particular defect are feed to the CNN so that discriminative features can be learned. To enhance the performance of its model they used data augmentation, simulating a variety of conditions and mitigating over-fitting. By doing so, the size of the dataset increases to millions of training samples for the model. However, and despite the good results, the model could not identify sub-classes, \eg fine roots from medium roots. 

J. Cheng and M. Wang use a fast regional convolutional neural network (fast R-CNN) to detect different classes of sewer defects and also to identify the coarse category to which they belong \cite{cheng2018automated}. Their model is comprised of a set of images gathered from video sewer inspections which are fed to the model to generate both classification and bounding box regression of the defect. Despite is implemented data augmentation, the similarities in the geometry of the sewers and color gradients and intensity, penalised the model performance.

So far, there are no studies that discuss the automated classification of sewer obstruction level using CNNs. Previous works focus on more general faulty elements in the sewer structure, \eg roots, cracks or deposits. However, due to the nature of the sewer system we work with (Barcelona area), it is crucial to assess if the sewer is free from obstacles, so that wastewater can flow through it ordinarily \cite{chataigner2020arsi}. That being said, we can still use some of the insights found when tackling similar tasks. 

\section{Sewer Data}
CETaqua is a public-private research institution dedicated to the design of more sustainable water management services. Along the years, CETaqua has gathered a database of videos from 6,590 human-made inspections made in the sewers of the area of Barcelona. Each video has an associated label, obtained from the operator’s report. The original distribution of videos is shown in Table \ref{tab:video_distribution}.

\begin{table}
    \begin{center}
    {\caption{Original video distribution.}\label{tab:video_distribution}}
    \begin{tabular}{ll}
    \hline
    Label & \# samples \\
    \hline
        clean          & 4720   \\
        slightly\_dirty  & 1146   \\ 
        dirty          & 235    \\ 
        very\_dirty    & 50 \\
        obstructed     & 98 \\
    \hline
    \hline
    Total:             & 6249 \\
    \hline  
    \end{tabular}
    \end{center}
\end{table}

\begin{figure}[b]
\centerline{\includegraphics[trim={0 0 0 .8cm},clip=True,width=0.49\textwidth]{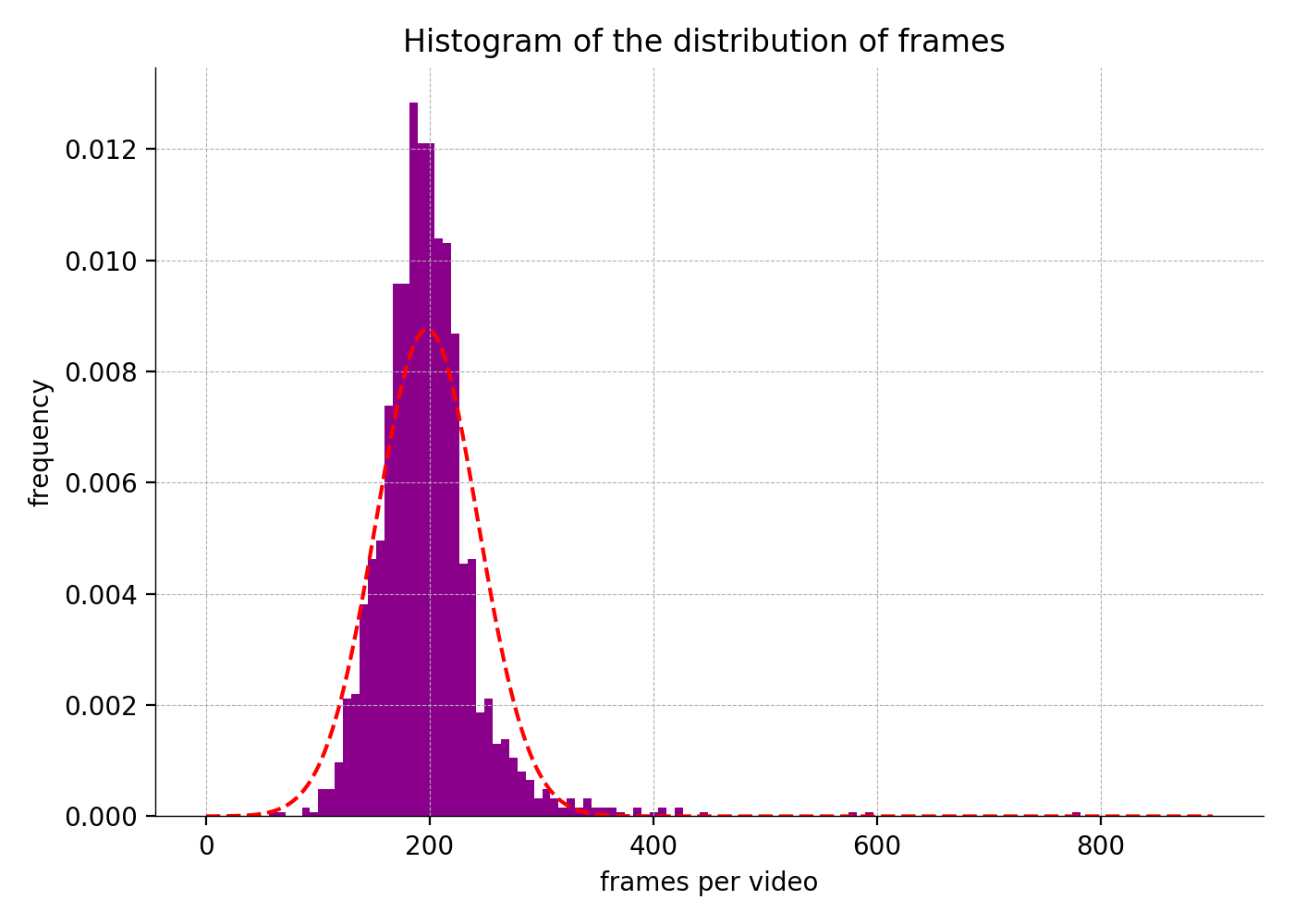}}
\caption{Distribution of the videos lengths.}
\label{fig:frames per video}
\end{figure}

Sewer videos were obtained by different operators following a shared set of guidelines: the operator brings the camera down into the sewer and starts recording the pipe from a static position. After a few seconds, the operator zooms in to look further into the end of the sewer, followed by a zoom out to the starting position. At that point, the video ends. Videos are mostly shot at 360x640 resolution, and 10Fps (frames per second). Prototypical sample frames for every obstruction level are shown in Figure \ref{fig:sewer_samples}. Since videos are recorded by different operators, there is a significant variance in their length. These range from \textasciitilde{18} to \textasciitilde{120} seconds. The distribution of video lengths is shown in Figure \ref{fig:frames per video}. No video is excluded from this study because of its length.

Notice the difference in number of samples between the most frequent and the least frequent classes is of two orders of magnitude. Such large class imbalance may handicap the learning process of many machine learning algorithms, including CNNs. To tackle this issue, and following the advise of use case industrial experts, we merge the two classes with less samples per class, "very dirty" and "obstructed", which are very close in meaning. We can do that without affecting the performance of the system because both classes imply the same industrial response once they are identified (\ie the prioritized cleaning of that particular sewer segment). 

After the merging, the number of elements in the minority class has increased (to 148), but the uneven data distribution remains relevant, which could lead to a severe bias in the model performance. To avoid that we balance the distribution of the data by randomly \textit{under-sampling} them to the minority class. Before data is fed into the model, we will still need to perform some pre-processing, to enable the learning process.

\section{Dataset Engineering}

The original task, as defined by the industrial requirements, is a video classification problem: Assign a given label to a given set of videos. However, we can reduce this to an aggregated image classification problem to simplify it, as the inherent temporal aspect of videos is mostly irrelevant for our case. Working with images also increases the number of training samples we can generate, as several frames from the same video become different (although not independent) training samples. With a larger training set we can improve the regularization and generalization of the CNN model.


Before transforming videos into images, we need to specify our dataset splits. Its essential to do so at this point, to avoid having images from the same video on both the train and test partitions. This would introduce a significant bias into the model, and significantly affect the relevance of our evaluation. After the \textit{under-sampling} process, the distribution of the videos is shown in table \ref{tab:video_distribution_2}. We have split the videos in two subsets: 70\% for the training process and the remaining 30\% to validate the model.

\begin{table}
    \begin{center}
    {\caption{Videos distribution per dataset split.}\label{tab:video_distribution_2}}
    \begin{tabular}{lll}
    \hline
    Label              & Train samples & Validation samples \\
    \hline
        clean          & 103 & 45     \\
        slightly\_dirty  & 103 & 45     \\ 
        dirty          & 104 & 44     \\ 
        very\_dirty    & 104 & 44     \\
    \hline
    \hline
    Total:             & 414 & 178    \\
    \hline
    \end{tabular}
    \end{center}
\end{table}

\subsection{Frame Selection}

\begin{figure}[b]
\centering
\includegraphics[width=0.49\textwidth]{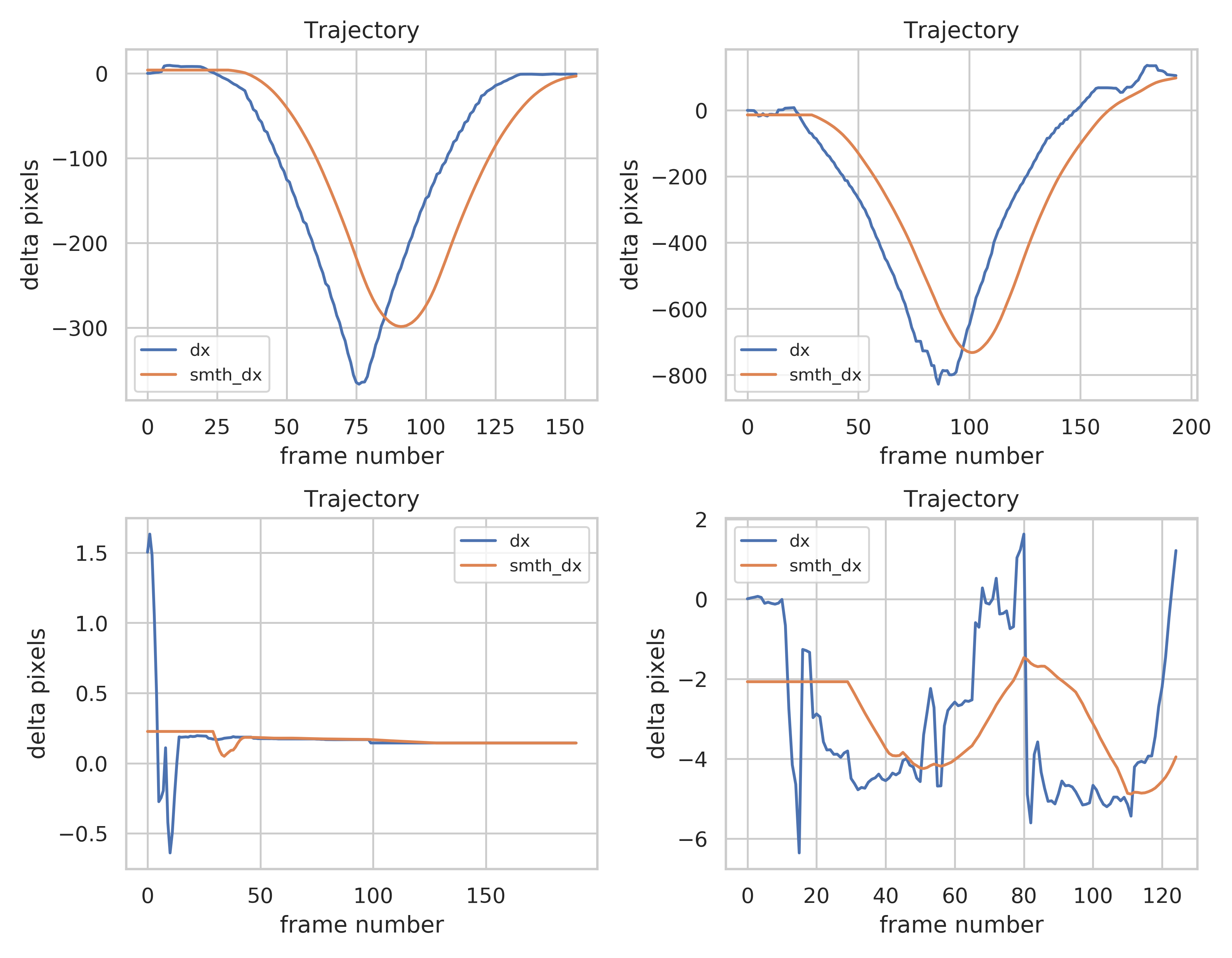}
\caption{Samples of pixels trajectories. Blue line shows change in pixel values. Red lines shows a smoothed version of the same function. Notice the significant scale variations in the vertical axis.}
\label{fig:smooth_traj}
\end{figure}

Of the full length of the video only a small portion of frames are usable for training. The zooming is digital on all cases, which means resolution is never increased, and some parts of the image are lost. For this reason, we gather the frames of the video where the camera is unzoomed. That is, from the beginning of the recording until the zooming in begins. To automatically locate this segment of interest, we used the VidStab video stabilization algorithm \footnote{http://nghiaho.com/?p=2093} from the OpenCV library \cite{opencv_library}. This algorithm produces a smoothed trajectory of pixels through the use of key point detectors. Figure \ref{fig:smooth_traj} shows examples of these smoothed trajectories.

The top row examples of Figure \ref{fig:smooth_traj} are prototypical videos, where the zoom in and zoom out stages form a clear 'V'. Unfortunately, not all videos are like that. There is a significant variance and noise in the extracted trajectories, as shown in the bottom row of Figure \ref{fig:smooth_traj}. Our analysis of trajectories shows that most videos have at least 3 seconds of image stability. Thus, we capture 30 consecutive frames for all videos. We do not extract a variable number of frames per video to avoid biasing the dataset. 

Using several frames per video results in the dataset distribution of Table \ref{tab:image_distribution}. Even though the number of images per class seems remarkable, this is deceptive. All images come from a hundred videos per class, which constrains significantly the variance of our training set. This also makes unproductive the use generalization techniques like data augmentation, since there are already plenty of similar images with small variations in our dataset.

\begin{table}
    \begin{center}
    {\caption{Images distribution per dataset split.}\label{tab:image_distribution}}
    \begin{tabular}{lll}
    \hline
    Label              & Train samples & Validation samples \\
    \hline
        clean          & 3090 & 1350     \\
        slightly\_dirty  & 3090 & 1350     \\ 
        dirty          & 3120 & 1320     \\ 
        very\_dirty    & 3120 & 1320     \\
    \hline
    \hline
    Total:             & 12420 & 5340    \\
    \hline
    \end{tabular}
    \end{center}
\end{table}

\subsection{Input Pipeline}

Most frames have a resolution of 360x640. They also have a vertical border as seen in Figure \ref{fig:sewer_samples}. After removing it, images are at 360x480 resolution, as seen in Figure \ref{fig:bad_classifications}. For those few images that had a slightly higher resolution, we applied a central crop. During our experimentation we noticed that models had the same performance if the 360x480 resolution was scaled down to 150x150. This is coherent with the task: since no specific object has to be identified, fine-grained detail is unnecessary. For this reason, our final training dataset is composed by 150x150 images. Resizing the images also reduced the number of parameters needed and the training costs (\ie time, power and money).

\section{Models}

CNNs models are composed by a sequence of stacked layers which learn increasingly complex representations from the data. For image inputs, these representations are visual abstractions of shape, patterns, colors \etc which are used as building blocks for perception. In the context of our problem, where the goal is to identify the amount of obstruction, complex patterns are irrelevant for the CNN. In other words, we do not care if the obstruction is caused by a bicycle or by a pile of cement. What is important to learn for the CNN is what a clean pipe looks like, and how different alterations to that normality correspond to different levels of obstruction. Clearly, spatial information is essential for the task, as sediment may be distributed along the channel, or it may form an obstruction at the bottom of the sewer. A sense of depth is also desirable, to assess obstructions proportions (and thus size) correctly. While we will not enforce these priors into the CNN, we will take them into account in our architectural designs, and we will validate them in our later interpretability study.

\subsection{Transfer Learning}

Fitting the many parameters found in deep CNNs to solve a task on high-dimensional inputs (\ie images) requires many data samples. To mitigate this need, one can use transfer learning: Initializing the parameters from a state optimized for a different problem, instead of initializing from a random state. Transfer learning is based on the assumption that most image challenges share a given set of visual properties which can be reused, instead of re-learnt. This is particularly true for low-level descriptors (\eg lines, angles \etc). Nevertheless, the transferability of features depends on the similarity between tasks, and in the variety and size of the data for which the pre-trained model was optimized \cite{yosinski2014transferable}. For this reason, the most popular source models for transfer learning are those containing a wide variety of patterns (\eg VGG16 \cite{simonyan2014very}) for a wide variety of goals (\eg ImageNet) \cite{azizpour2015factors}.

Considering the limited number of samples available in our task (remember images come from a small set of videos), in this section we consider transfer learning as a potentially useful approach. We explore this hypothesis by using a VGG16 architecture trained on the ImageNet dataset. To adjust the VGG16 model to our needs, we start by removing the parameters of the original classifier (\ie the two fully-connected layers), since these are too optimized for the original problem. We also addapt the output of the network (originally, a 1,000 classes problem) to fit our task. With this setting in place, we can now train the network through fine-tuning.

When fine-tuning, one must decide which layers to freeze (\ie fixing the weights), which to re-train (\ie fine-tuning the weights) and which to replace (\ie randomly initialized) or delete. The more layers we freeze, the more similar both tasks should be. Unfortunately, our industrial case is a rather unique one, even when compared with a broad classification task like ImageNet. In our experiments, we tried freezing a variable number of convolutional layers gradually bottom up (remember, the fully-connected layers are always randomly initialized and thus always replaced and optimized). Significantly, none of these experiments were successful. In all experiments, the model either overfitted to the data or failed to learn meaningful representations. We hypothesise that the particularity of our problem makes it hard to reusing patterns learned on general purpose datasets. Indeed, there is little in common between discriminating dog breeds and computing the level of obstruction of a sewer. On the other hand, the huge number of parameters in networks trained for large tasks like ImageNet is inadequate for a small problem like ours. 

\subsection{Architecture proposed}

Since transfer learning was unsuccessful, we decided to define an architecture design top to bottom for the sewer classification problem. This is motivated by the uniqueness of our problem. We started from a shallow architecture, and increased its size until underfitting was no longer an issue. At that point, we optimized the hyper-parameters to get the best working model. The Table \ref{tab:cnn_small} shows the final CNN design.

\begin{table}
\begin{center}{\caption{CNN architecture proposed.}\label{tab:cnn_small}}
\begin{tabular}{lll}
\hline
                 Layer (type) &          Output Shape &   \# Param  \\
\hline
               conv1 (Conv2D) &  (150, 150, 32) &       896 \\
         pool1 (MaxPooling2D) &    (75, 75, 32) &         0 \\
               conv2 (Conv2D) &    (75, 75, 32) &      9248 \\
         pool2 (MaxPooling2D) &    (38, 38, 32) &         0 \\
               conv3 (Conv2D) &    (38, 38, 64) &     18496 \\
         pool3 (MaxPooling2D) &    (19, 19, 64) &         0 \\
            flatten (Flatten) &         (23104) &         0 \\
                  fc1 (Dense) &          (1024) &  23659520 \\
           dropout1 (Dropout) &          (1024) &         0 \\
               logits (Dense) &             (4) &      4100 \\
\hline
\hline
     Total params: 23,692,260 &                       &           \\
 Trainable params: 23,692,260 &                       &           \\
\hline
\end{tabular}
\end{center}
\end{table}

Notice the relatively small size of the architecture. Increasing the number of filters and the kernel size provided no improvement, mostly because the variety of patterns to learn is small: the model does not have to recognize all possible objects and shapes that may obstruct the sewer. It must limit itself to learn what a clean sewer looks like, and what obstructions represent visually in that regard. Coherently, our experiments shows that a CNN with only three convolutional layers yields the best results. On top of that we add a fully-connected layer, accounting for 99.9\% of the parameters of the CNN, to learn to discriminate between the different levels of obstruction. The size of this last layer was also optimized empirically. In our experiments we used the Cross-Entropy loss function, ADAM optimizer with a \num{1e-6} learning rate, and 0.5 dropout value.


\section{Evaluation and Results}

\begin{figure}[tbh]
\centerline{\includegraphics[width=0.49\textwidth]{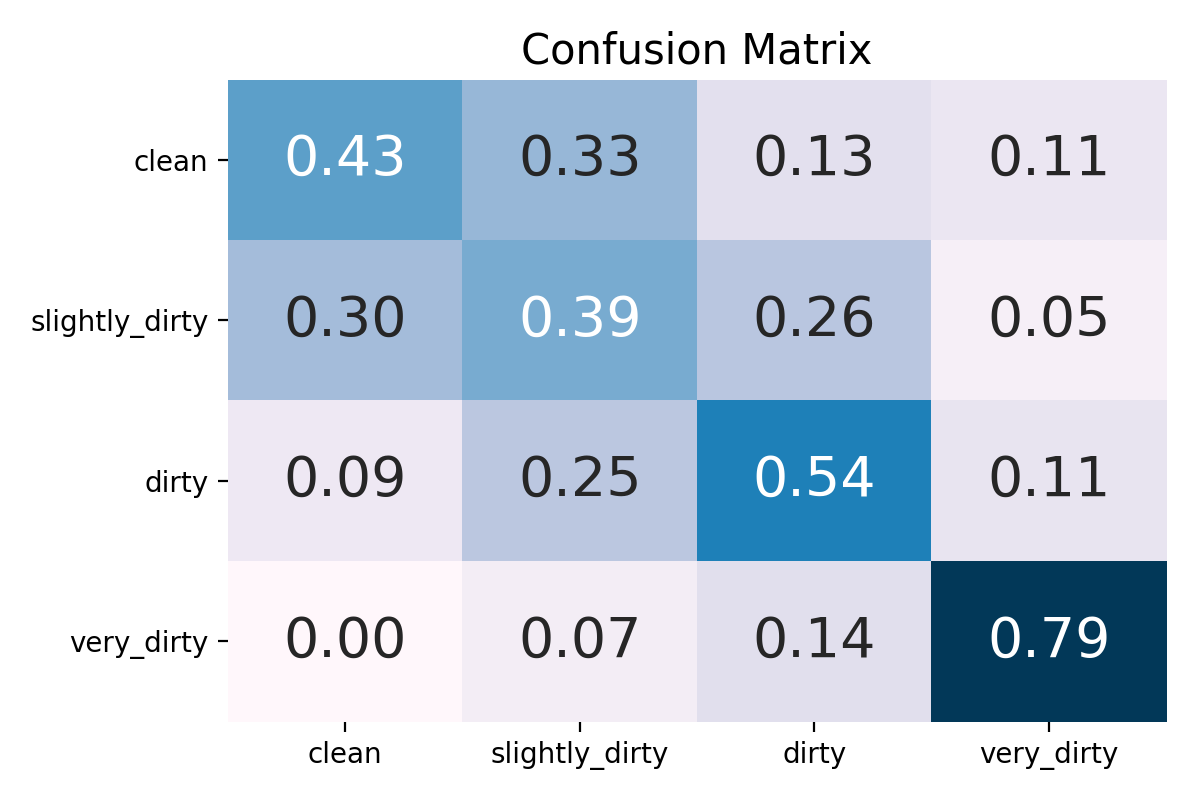}}
\caption{Normalized image-wise confusion matrix for validation set.}
\label{fig:cm}
\end{figure}

To evaluate the performance of the trained model, we start by the confusion matrix. This will allow us to understand the frequency and severity of the mistakes being made by the model. As shown in Figure \ref{fig:cm}, 53.7\% of images are classified in the correct class. 34.7\% of images are classified in a neighboring class (\eg slightly dirty as dirty). The fact that mistakes are centered around the diagonal indicates that the model is properly learning the nature of the problem. It is also worth noticing how the most relevant classes for the industrial application (the dirty and very dirty ones) are the ones classified with the highest accuracy.

\begin{figure}[tbh]
    
    
        \centerline{\includegraphics[width=0.49\textwidth]{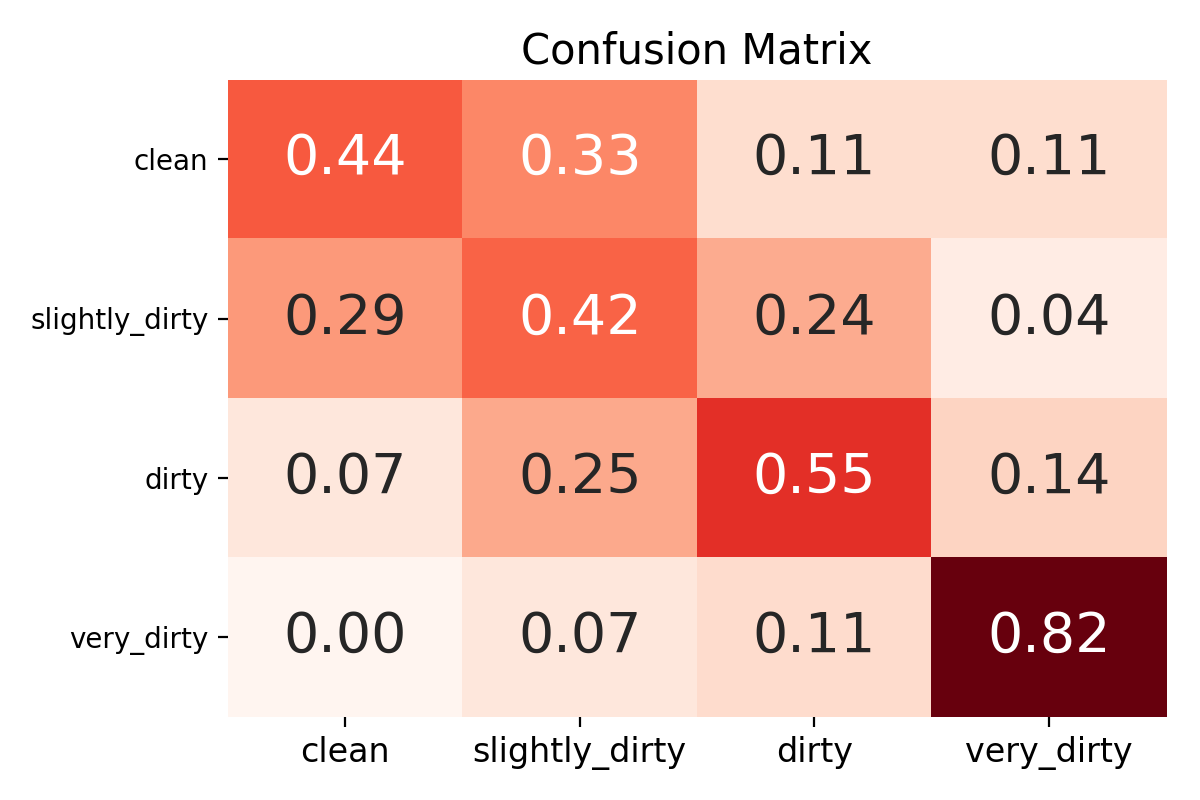}}
        \caption{normalized video-wise confusion matrix on the validation set. From the voting of classified images.}
        \label{fig:cm_videos}

\end{figure}

The previous metric was computed image-wise, in the context of an image classification task. However, our final purpose is to provide a video classification tool. Based on the CNN image predictions, we generate a video classifier using a voting strategy, where each image from a video contributes with one vote towards the classification of the video itself. The confusion matrix of Figure \ref{fig:cm_videos} shows the video-wise classification results. In this case, 55.7\% of images are classified in the correct class, 2\% more than the image classifier. The images classified in a neighboring class decrease 0.7\%, to 34.0\%. The performance of this model fits the requirements of the industrial task, and is already profitable from a practical point of view.

\begin{figure}[tb]
\centerline{\includegraphics[width=0.49\textwidth]{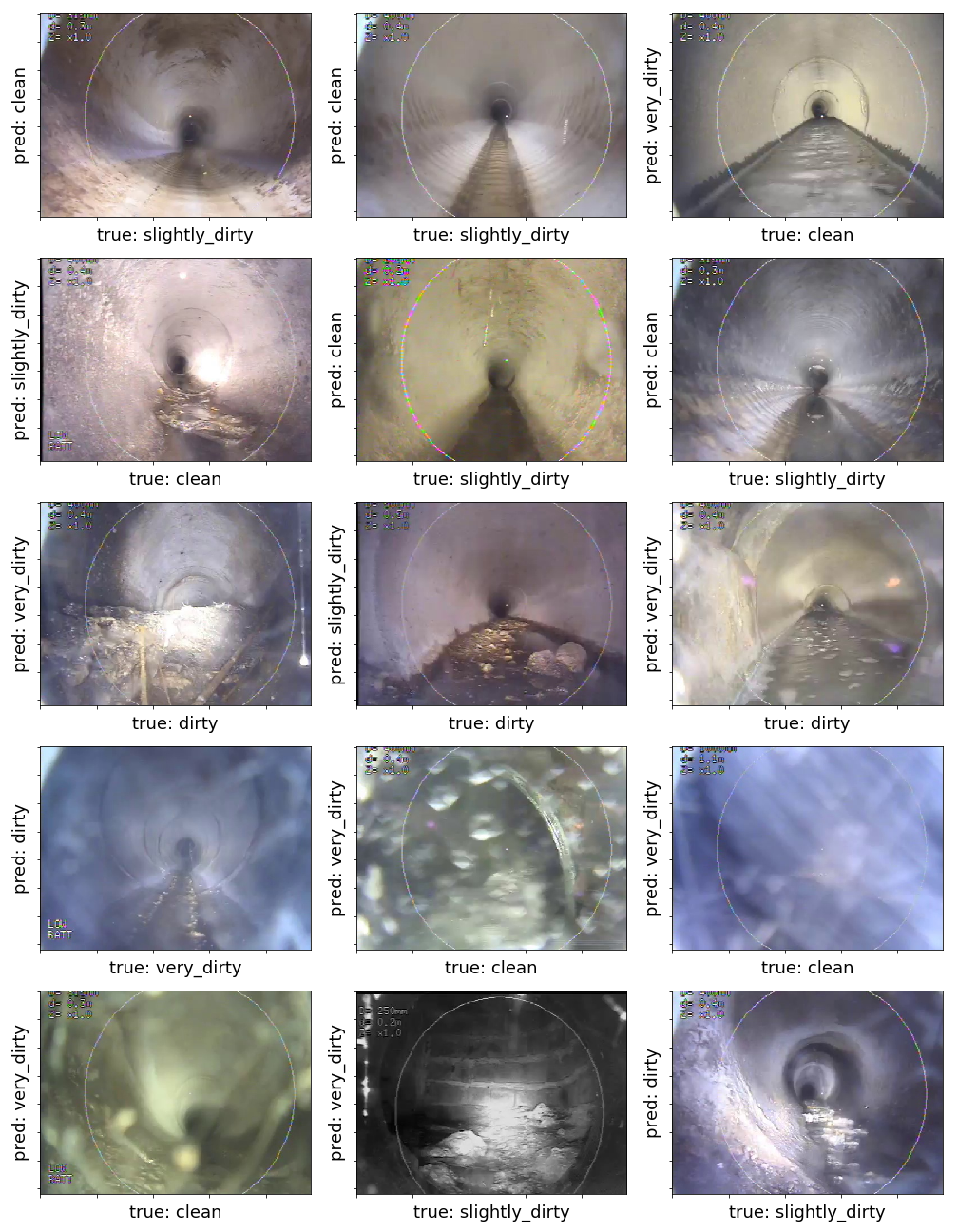}}
\caption{Samples of miss-predictions. \textit{true} indicates ground truth. \textit{pred} indicates model prediction.}
\label{fig:bad_classifications}
\end{figure}

Beyond the numeric analysis of the classifier outcome, we also explore the mistakes done by the model. Figure \ref{fig:bad_classifications} presents some representative examples of failed predictions. The first two rows contains examples of videos where the labeling seems to be erroneous, which we attribute to human error. These samples could be re-labeled to improve the training dataset quality and the model performance. The third row shows examples where the labeling criteria seems to be inconsistent, as a result of having multiple operators labeling videos. Although the model predictions in these cases count as miss-classifications, its criteria seems adequately coherent. Another complicating factor we found in our analysis of mistakes is rain. Examples of these are shown in the forth row of Figure \ref{fig:bad_classifications}. Rain introduces lots of noise in the images, which handicaps perception and model prediction. Finally, the last row shows cases where the perspective of the camera is not normative (\ie centered in the pipe and looking towards the end of it). These variations confuse the model. To bypass this limitation more training data is needed.

\subsection{Interpretability}

So far we have gathered evidence that the model is learning properly. Nevertheless, trusting the predictions of a black-box is never ideal, no matter how confident we are on its performance. Explainability of the model is crucial for industrial risk assessment and regulation compliance. Thus, we take one more step into the validation of the model by looking at the visual patterns learned and used by the model to classify the data. This will provide interpretability to our system.

\begin{figure}[ht]
    \centering
    \begin{subfigure}[b]{0.49\textwidth}
        \centerline{\includegraphics[trim={2.5cm 24.5cm 2.5cm 3.7cm},clip=True,width=1\textwidth]{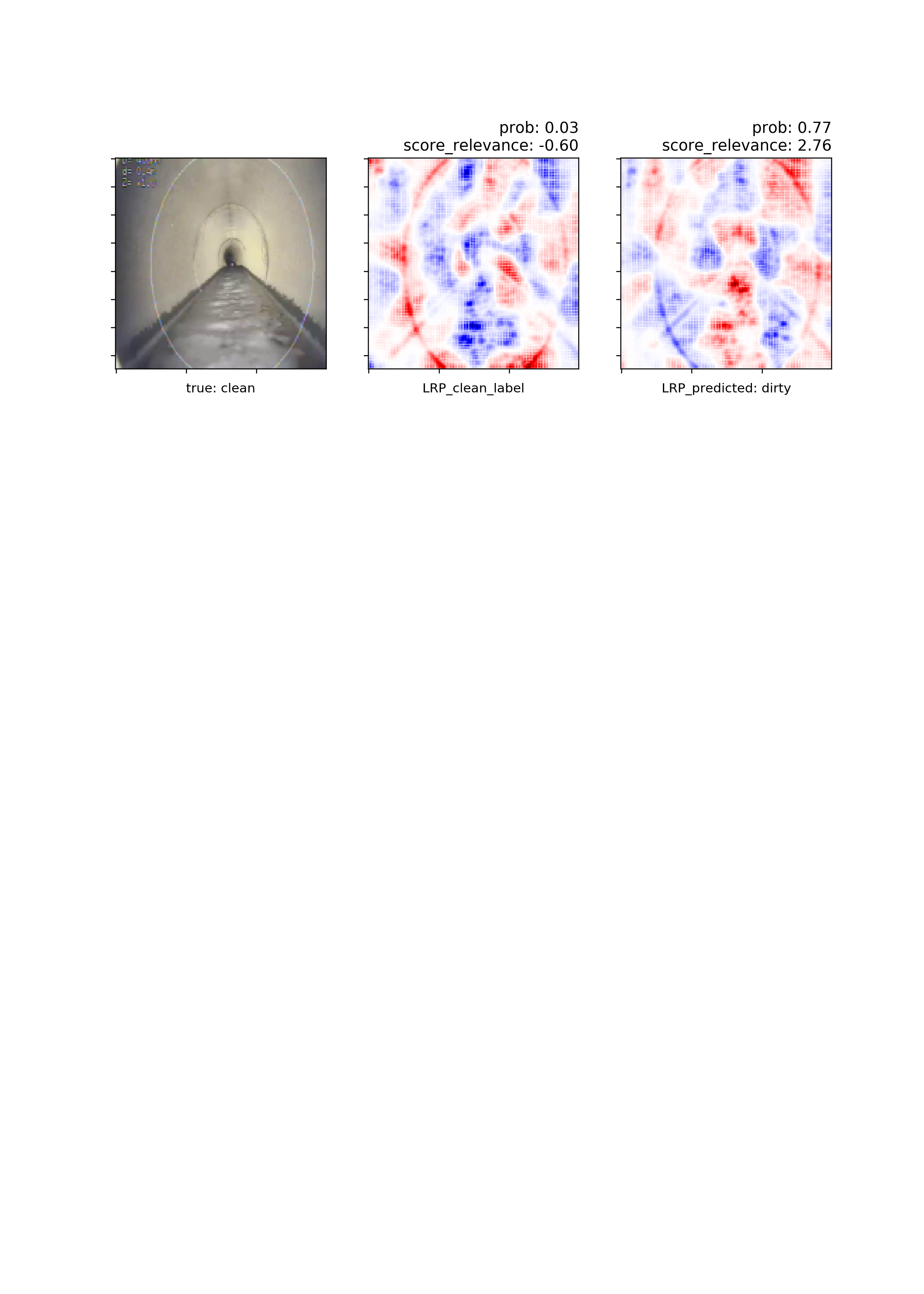}}
    \end{subfigure}
    \hfill
    \begin{subfigure}[b]{0.49\textwidth}
        \centerline{\includegraphics[trim={2.5cm 24.5cm 2.5cm 3.7cm},clip=True,width=1\textwidth]{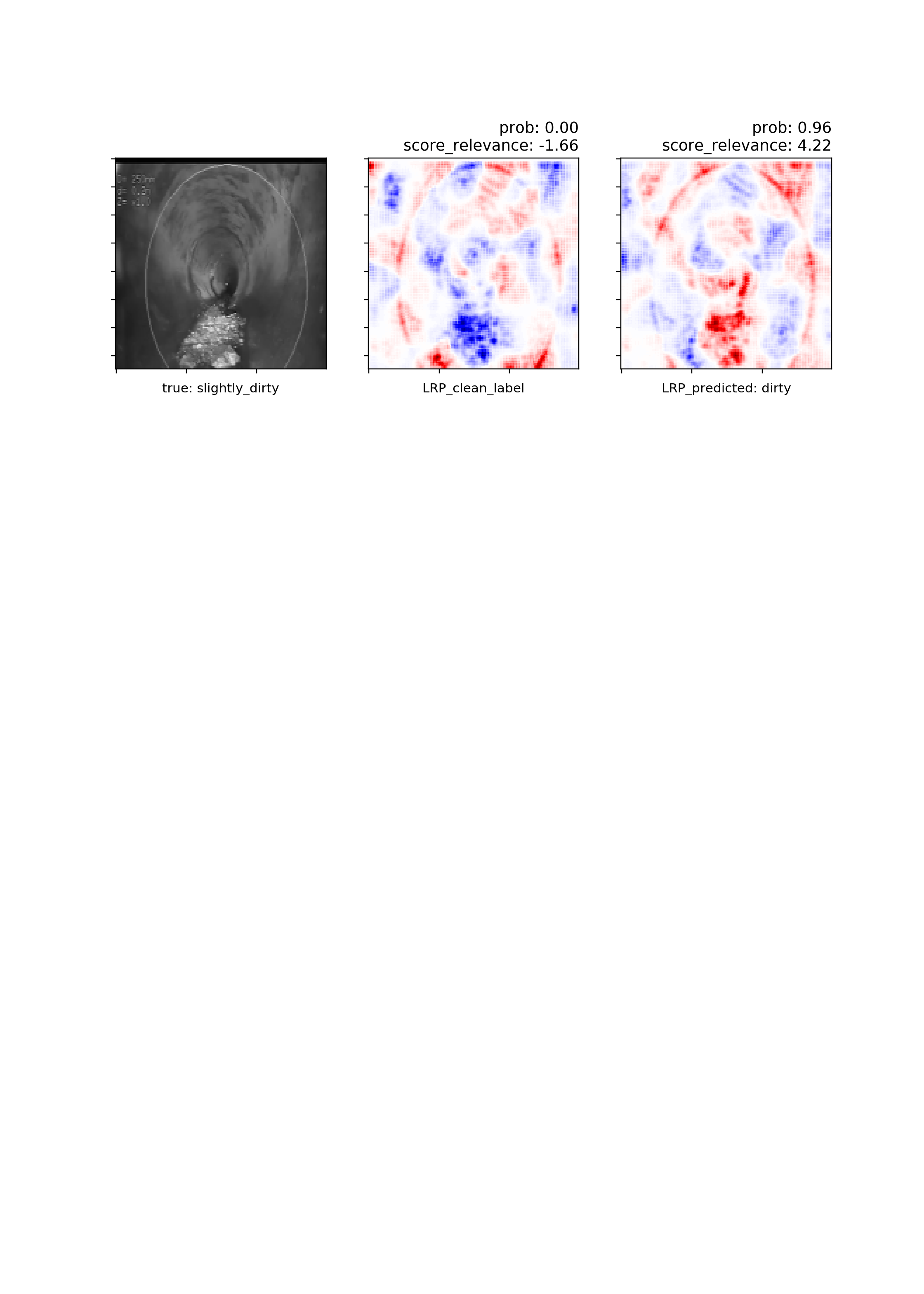}}
    \end{subfigure}
    \hfill
    \begin{subfigure}[b]{0.49\textwidth}
        \centerline{\includegraphics[trim={2.5cm 24.5cm 2.5cm 3.7cm},clip=True,width=1\textwidth]{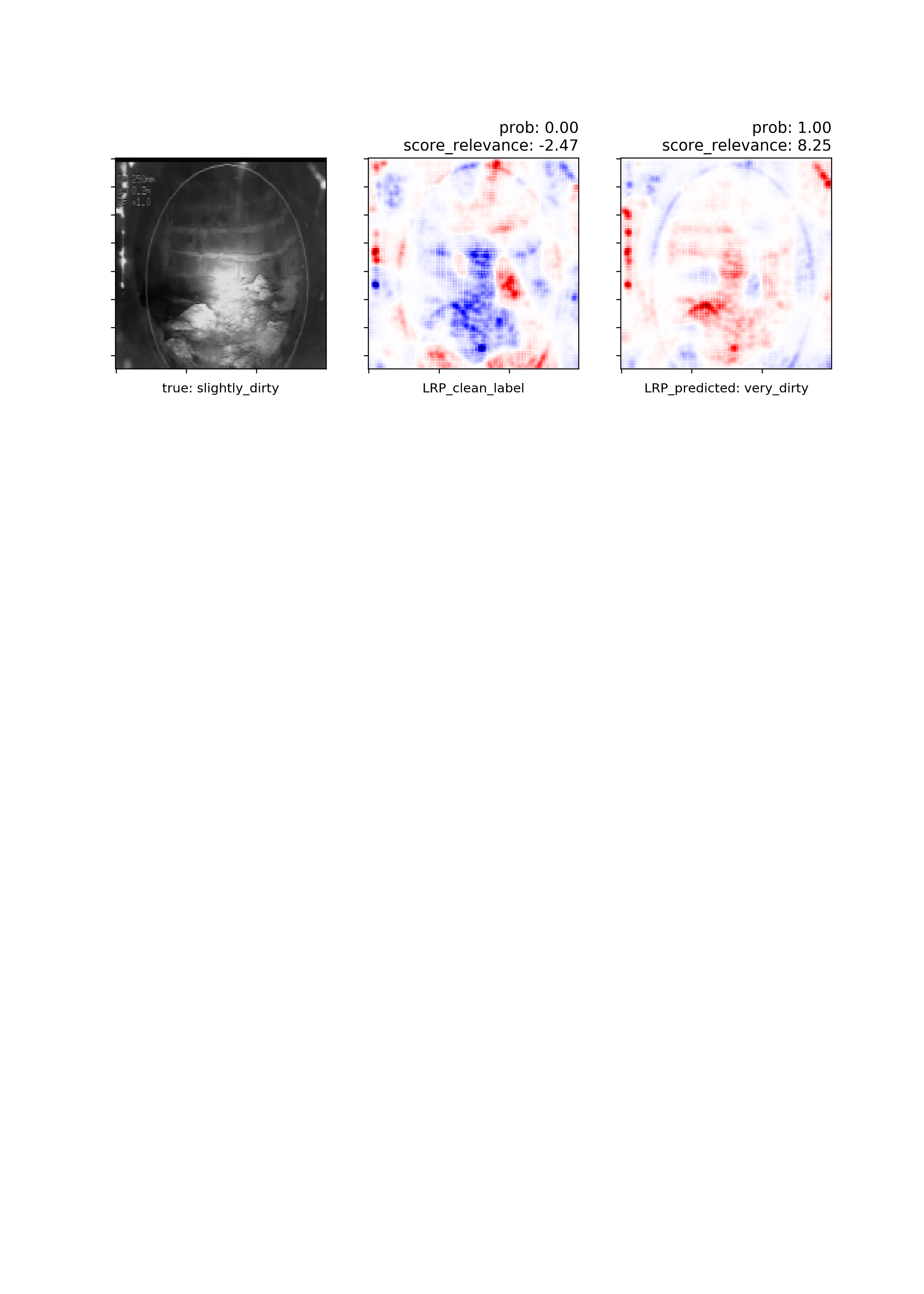}}
    \end{subfigure}
    \hfill
    \begin{subfigure}[b]{0.49\textwidth}
        \centerline{\includegraphics[trim={2.5cm 24.5cm 2.5cm 3.7cm},clip=True,width=1\textwidth]{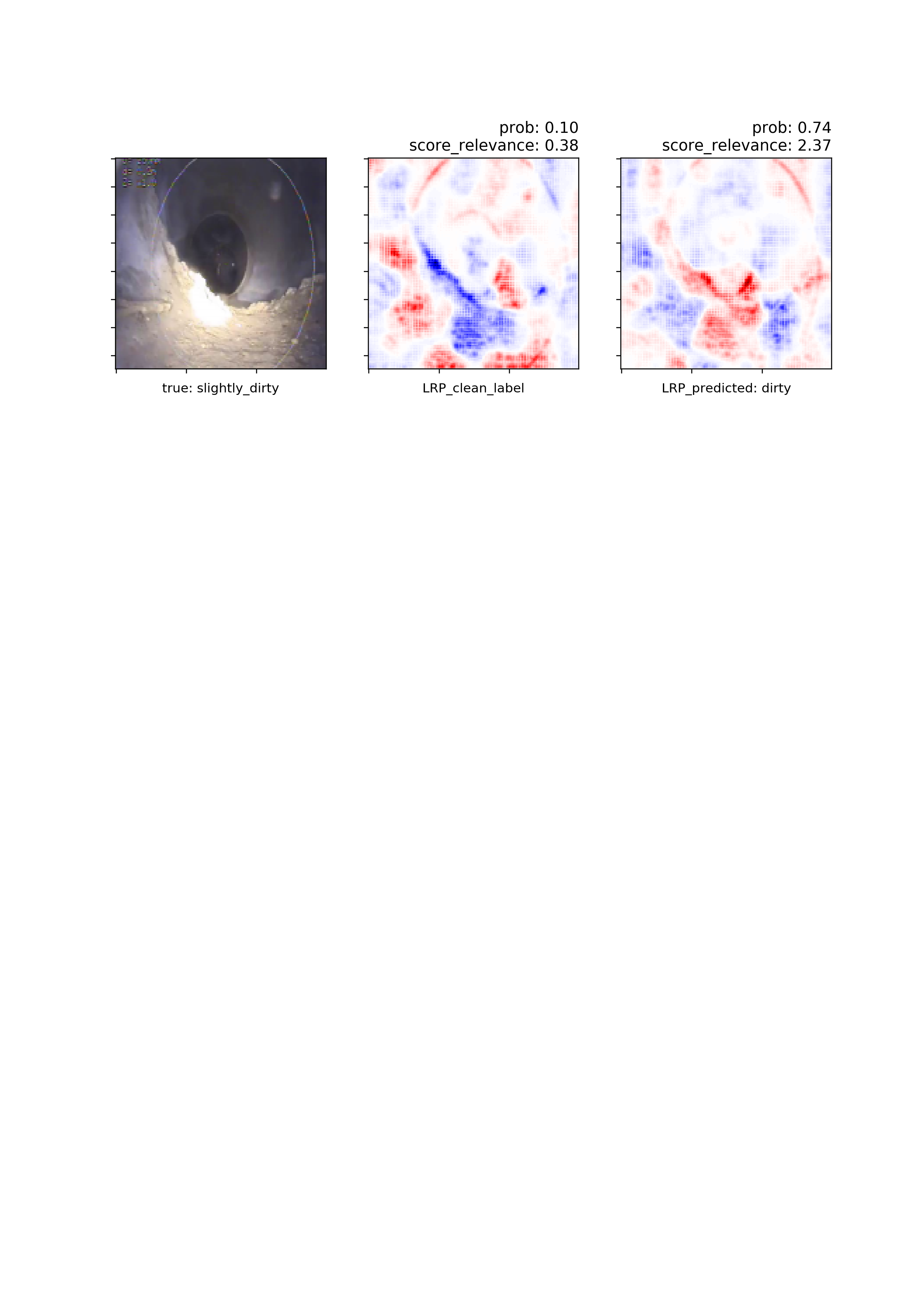}}
    \end{subfigure}
    \hfill
    \begin{subfigure}[b]{0.49\textwidth}
        \centerline{\includegraphics[trim={2.5cm 24.5cm 2.5cm 3.7cm},clip=True,width=1\textwidth]{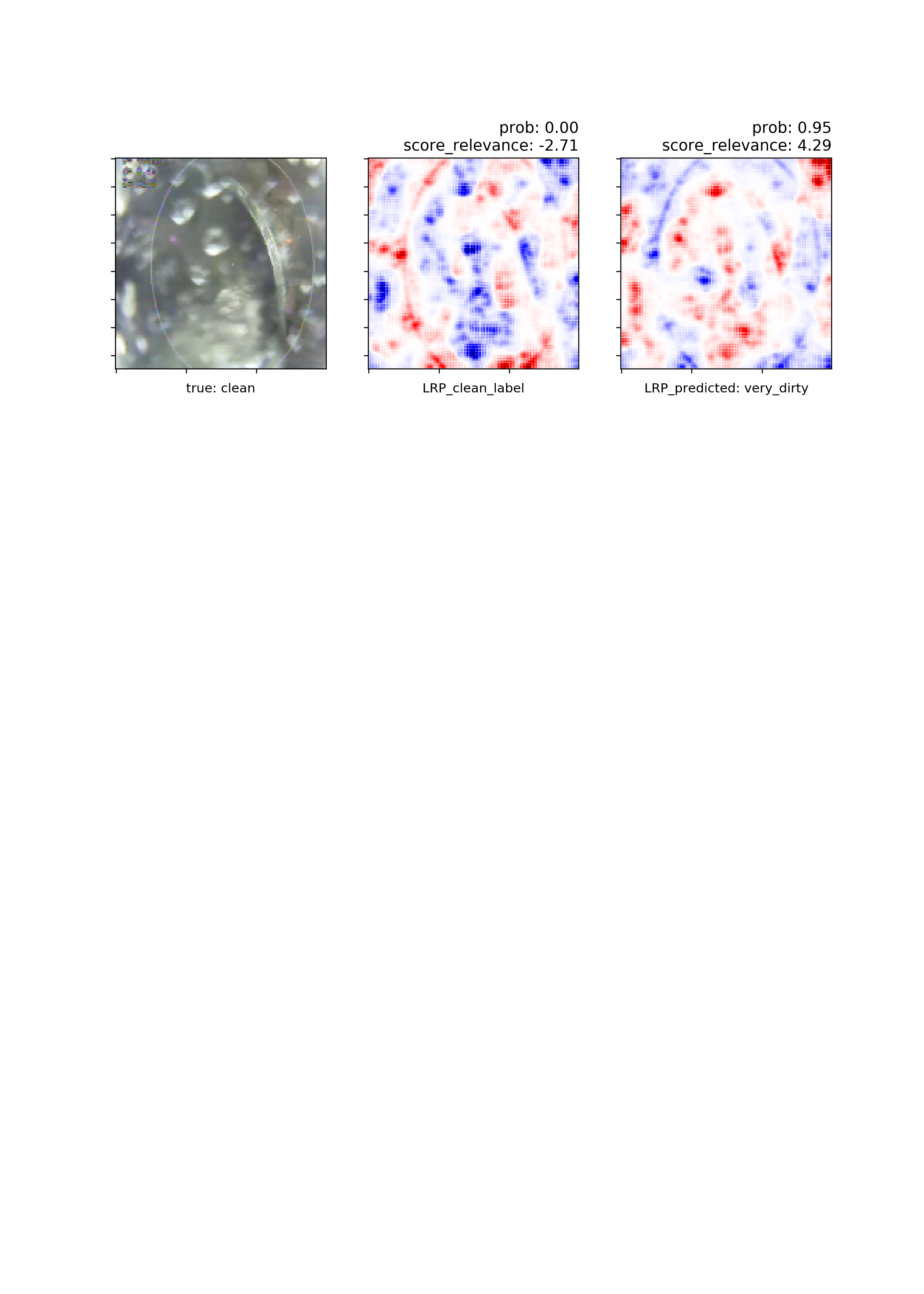}}
    \end{subfigure}
        \caption{Each row shows an independent example. First column contains the original image. Second and third columns show the LRP of clean and dirty labels, together with the confidence of the prediction (score relevance). Red pixels indicate evidence in favor of prediction, blue pixels indicate evidence against it. LRP plots show values at different scales.}
        \label{fig:lrp_bad_classifications}
\end{figure}

The Layer-wise Relevance Propagation (LRP) is presented in Bach \etal \cite{bach2015pixel}. This algorithm works on a trained model, trying to identify which features of the input image have the highest relevance for the prediction of that image. Relevance is backpropagated from the output layer, assigning scores to the application of features, layer by layer until reaching the input. Each layer stores an equal amount of relevance, which is variably distributed among its features. The relevance of pixels in the input can be conveniently visualized through heatmaps. We integrate the LRP to our trained model, to explore its decision making process. A sample of the result can be seen in Figure \ref{fig:lrp_bad_classifications}.


For visibility reasons, LRP values are not normalized among all plots (\ie the same color on different LRP may indicate different relevance). The reference value for each LRP plot is shown above it (score relevance), and it depends on the confidence of that particular prediction. If all colors were normalized, colors from predictions with lower confidence would be barely visible. For this reason, plots of low probability predictions should not be over-interpreted.

Let us first consider what evidence is used to predict obstructions. As shown in the second column of Figure \ref{fig:lrp_bad_classifications}, the main evidence used for justifying a high level of obstruction (\ie \textit{dirty} or \textit{very dirty}) is located at the ground along the pipe. This seems adequate, since this center canal will be naturally occupied by most obstructions. The LRP plots also indicate that changes in illumination are taken as evidence of obstruction (\eg third row, third column). Coherently, in a clean pipe light is smoothly distributed, while obstructed pipes contain segments of extreme illumination contrast.

On the other hand, clean predictions seem to focus on circular shapes. These shapes are periodic within pipes, and their visibility is used by the CNN as evidence of cleanness. Among the circles the CNN is locating, we find an artificial guidance circle introduced to help the operator center the camera within the pipe. This visual aid is also being used by the CNN for prediction. With the current results we are unable to assert if the classifier would perform better without the visual aid or not. Regarding the use of circles for clean predictions, we find this a quite consistent policy, as any large obstruction would occlude these circles (in the cases of pipe circles), or would make them invisible due to changes in the illumination (in the case of the visual aid circle, as in the forth row).

The use of both the ground path and illumination contrast as features for prediction explains the difficulties of the model for predicting images where there is either rain or changes in perspective. As shown in the bottom three rows of Figure \ref{fig:lrp_bad_classifications}, the model still focuses on these features, even though in these cases such features characterize noise instead of obstructions.



\section{Industrial Deployment}\label{Industrial_Deployment}

Our purpose is to help improve the efficiency of cleaning operations. We propose to do so through an automated mechanism for the evaluation of sewer conditions, fueled by the CNN model previously described. In this section, we outline the rest of the necessary components for the implementation of the solution in the real environment. We design it so that the system keeps learning once deployed. The two main system components: a labeler API and a training pipeline.

The labeler API is to be integrated into the IT systems of the maintenance department. It provides both automatic classification of videos, as well as a labeling interface for humans. Once a new video inspection is uploaded, the API is automatically requested for a classification. This will be done on a number of random frames from the static part of the video. The result, both classification and confidence, is processed by a rule-based system. This determines what to do with the video. Rules such as: If the classification is \textit{dirty} or \textit{very dirty} and the confidence is high, send it to the cleaning team with urgency. If the confidence is low, send it to the queue for human labeling. If classification is \textit{slightly dirty} or \textit{clean} with high confidence, send it to the queue with low priority. All video labeled by humans through the interface are automatically used in the training pipeline.

The training pipeline is defined in a continuous integration server. When new videos arrive, these are fed into an object storage server. The storage server can triggers a series of jobs, after a minimum number of new samples are received. These jobs execute the following pipeline:
\begin{enumerate}
    \item Dataset balancing and split
    \item Video stabilization and frame selection 
    \item Frame resizing 
    \item Model training
    \item Model evaluation
\end{enumerate}

The result of this pipeline is a model in \textit{TensorFlow} along with a PDF document containing a sample of automatically labeled frames that are to be reviewed by an expert. If, according to this expert, the results are good --\ie the classification of sewer images is adequate-- the model is automatically deployed to the production API server, replacing the previous version. Every pipeline that generated every version of the model is stored along with the data used in it. This provides full reproducibility to the system.

The system is designed for low-degree maintenance. Together with a heavy automatization we propose to scale resources to the cloud, accessing GPU resources only when training. That is periodically and for less than an hour. It is also designed for re-usability. The same pipeline could be potentially applied to any sewer system that shares strong similarities --both structural and sedimental-- with the one we have worked with. If differences were significant the CNN model architecture should be reassessed. It is therefore our assumption that this solution could be deployed internationally to any sewer management that uses video sampling inspections.
 
Beyond technical contributions, several improvements could be made to the process. For example, using higher resolution cameras, adding stabilization gear, giving more specific instructions to operators or applying heavier pre-processing techniques. However, relying on such improvements would reduce the generalization power of the model. Low-quality data, something frequent in sewer inspections, is something to be learned. Thus, we consider our current approach --dealing with our current datasets, as faulty as they might be-- more beneficial for the industrial purpose in the long term.

\section{Conclusions}

The proper operation of sewers is critical for current societies: It conveys domestic sewage, industrial wastewater, and storm-water runoff. The efficient and scalable identification of obstructions in sewer infrastructure is critical for their correct maintenance, given the sewer network length and the up-time requirements of the service. In this context, operators are put under severe pressure, forced to quickly record and evaluate inspections daily.

In this work, we seek to alleviate the pressure on human performance through a CNN model trained to identify the level of obstruction of a sewer. We start by reducing the problem to an image classification one, as this is a more scalable and constrained approach. A pixel motion analysis allows us to measure the degree of noise in the dataset (which is high), and to define a unified frame extraction policy. Given a significant imbalance among target classes, we are forced to merge two similar classes and to down-sample the rest. In this setting we perform our experiments.

In sight of the limited data availability, we decide to first use transfer learning, exploiting features learned for a different problem. This approach failed, most likely, due to the dissimilarities between tasks. Unfortunately this is a recurrent issue in industrial domains, where data follows a very particular distribution, with little in common with large, popular datasets. It remains to be seen if more flexible transfer learning mechanisms, like feature extraction where it is not needed to re-train the CNN \cite{garcia2018out}, would be feasible in this setting. 

In our experiments the best results are obtained by a rather small and shallow architecture, consistently with the nature of the task: There is no need to learn any specific pattern, just an overall sense of space and obstruction. The evaluation indicates this model learns to solve the task satisfactorily, and illustrates the main reasons behind the failed predictions. Most frequently, inconsistent human labeling, variations in perspective and environmental noise like rain.

We explore the behavior of the model by looking at the relevance of input pixels for output classification. This allows us to validate the visual features used by the model to make predictions. In particular, we notice how the center canal of the sewer is essential for the assessment of obstructions, how the visibility of circles around the pipe speaks for cleanness, and how changes in illumination and perspective can complicate the resolution of the problem. Based on these, we formulate two recommendations for current inspection protocols: to pay special attention to camera location before starting the recording and to avoid doing inspections under heavy rain.

Two more complicating factors were identified in the data during the development work. First, human mistakes when labeling videos. These are unfortunately frequent and bias the performance measures obtained for the model. In other words, the model may be predicting better than what is measured. Second, the variability in labeling criteria. This is one of the motivating factors of this work, as an unstable policy reduces the quality and efficiency of maintenance interventions. All experimental outcomes suggest that the trained model has a more consistent behavior than human labeling. This already makes the solution appealing from an industrial perspective.

Beyond the visual model, we propose an integral system design to deploy all desirable functionalities. This includes an API, through which videos can be automatically labeled by the model, while also providing a common interface for human labeling. It also includes a training pipeline, so that models can be periodically trained and deployed in production with minimal effort. 

To generate the deployment model we will retrain the system using all available data (\ie including the validation set). Before that, we will try to reduce dataset noise by fixing some of the most obvious labeling mistakes, as well as removing videos with rain. For this last case, and while data availability remains limited, we consider best to train a model to discriminate between images with rain and images without rain, so that the system can automatically inhibit itself in favor of humans when asked to classify videos in rainy conditions.




    


\ack This work is partially supported by the Consejo Nacional de Ciencia y Tecnologia, No. CVU: 630716, by the RIS3CAT Utilities 4.0 SENIX project (COMRDI16-1-0055), cofounded by the European Regional Development Fund (FEDER) under the FEDER Catalonia Operative Programme 2014-2020. It is also partially supported by the Spanish Government through Programa Severo Ochoa (SEV-2015-0493), by the Spanish Ministry of Science and Technology through TIN2015-65316-P project, and by the Generalitat de Catalunya (contracts 2017-SGR-1414).

\bibliography{ecai}

\begin{thebibliography}{10}

\bibitem{abdel2003analysis}
Ikhlas Abdel-Qader, Osama Abudayyeh, and Michael~E Kelly, `Analysis of
  edge-detection techniques for crack identification in bridges', {\em Journal
  of Computing in Civil Engineering}, {\bf 17}(4),  255--263, (2003).

\bibitem{azizpour2015factors}
Hossein Azizpour, Ali~Sharif Razavian, Josephine Sullivan, Atsuto Maki, and
  Stefan Carlsson, `Factors of transferability for a generic convnet
  representation', {\em IEEE transactions on pattern analysis and machine
  intelligence}, {\bf 38}(9),  1790--1802, (2015).

\bibitem{bach2015pixel}
Sebastian Bach, Alexander Binder, Gr{\'e}goire Montavon, Frederick Klauschen,
  Klaus-Robert M{\"u}ller, and Wojciech Samek, `On pixel-wise explanations for
  non-linear classifier decisions by layer-wise relevance propagation', {\em
  PloS one}, {\bf 10}(7),  e0130140, (2015).

\bibitem{belles2015kinect}
Crist{\'o}bal Bell{\'e}s and Filiberto Pla*, `A kinect-based system for 3d
  reconstruction of sewer manholes', {\em Computer-Aided Civil and
  Infrastructure Engineering}, {\bf 30}(11),  906--917, (2015).

\bibitem{opencv_library}
G.~Bradski, `{The OpenCV Library}', {\em Dr. Dobb's Journal of Software Tools},
  (2000).

\bibitem{cha2017deep}
Young-Jin Cha, Wooram Choi, and Oral B{\"u}y{\"u}k{\"o}zt{\"u}rk, `Deep
  learning-based crack damage detection using convolutional neural networks',
  {\em Computer-Aided Civil and Infrastructure Engineering}, {\bf 32}(5),
  361--378, (2017).

\bibitem{chataigner2020arsi}
Fran{\c{c}}ois Chataigner, Pedro Cavestany, Marcel Soler, Carlos Rizzo,
  Jesus-Pablo Gonzalez, Carles Bosch, Jaume Gibert, Antonio Torrente, Ra{\'u}l
  Gomez, and Daniel Serrano, `Arsi: An aerial robot for sewer inspection', in
  {\em Advances in Robotics Research: From Lab to Market},  249--274, Springer,
  (2020).

\bibitem{cheng2018automated}
Jack~CP Cheng and Mingzhu Wang, `Automated detection of sewer pipe defects in
  closed-circuit television images using deep learning techniques', {\em
  Automation in Construction}, {\bf 95},  155--171, (2018).

\bibitem{dang2018utilizing}
L~Minh Dang, Syed~Ibrahim Hassan, Suhyeon Im, Irfan Mehmood, and Hyeonjoon
  Moon, `Utilizing text recognition for the defects extraction in sewers cctv
  inspection videos', {\em Computers in Industry}, {\bf 99},  96--109, (2018).

\bibitem{donat2017addendum}
Markus~G Donat, Andrew~L Lowry, Lisa~V Alexander, Paul~A O'Gorman, and Nicola
  Maher, `Addendum: More extreme precipitation in the world's dry and wet
  regions', {\em Nature Climate Change}, {\bf 7}(2),  154, (2017).

\bibitem{esquivel2009reconstruction}
Sandro Esquivel, Reinhard Koch, and Heino Rehse, `Reconstruction of sewer shaft
  profiles from fisheye-lens camera images', in {\em Joint Pattern Recognition
  Symposium}, pp. 332--341. Springer, (2009).

\bibitem{garcia2018out}
Dario Garcia-Gasulla, Armand Vilalta, Ferran Par{\'e}s, Eduard Ayguad{\'e},
  Jesus Labarta, Ulises Cort{\'e}s, and Toyotaro Suzumura, `An out-of-the-box
  full-network embedding for convolutional neural networks', in {\em 2018 IEEE
  International Conference on Big Knowledge (ICBK)}, pp. 168--175. IEEE,
  (2018).

\bibitem{halfawy2013efficient}
Mahmoud~R Halfawy and Jantira Hengmeechai, `Efficient algorithm for crack
  detection in sewer images from closed-circuit television inspections', {\em
  Journal of Infrastructure Systems}, {\bf 20}(2),  04013014, (2013).

\bibitem{huynh20163d}
Phat Huynh, Robert Ross, Andrew Martchenko, and John Devlin, `3d anomaly
  inspection system for sewer pipes using stereo vision and novel image
  processing', in {\em 2016 IEEE 11th Conference on Industrial Electronics and
  Applications (ICIEA)}, pp. 988--993. IEEE, (2016).

\bibitem{iyer2006segmentation}
Shivprakash Iyer and Sunil~K Sinha, `Segmentation of pipe images for crack
  detection in buried sewers', {\em Computer-Aided Civil and Infrastructure
  Engineering}, {\bf 21}(6),  395--410, (2006).

\bibitem{kumar2018automated}
Srinath~S Kumar, Dulcy~M Abraham, Mohammad~R Jahanshahi, Tom Iseley, and Justin
  Starr, `Automated defect classification in sewer closed circuit television
  inspections using deep convolutional neural networks', {\em Automation in
  Construction}, {\bf 91},  273--283, (2018).

\bibitem{lattanzi20143d}
David Lattanzi and Gregory~R Miller, `3d scene reconstruction for robotic
  bridge inspection', {\em Journal of Infrastructure Systems}, {\bf 21}(2),
  04014041, (2014).

\bibitem{montero2015past}
Roberto Montero, Juan~G Victores, Santiago Martinez, Alberto Jard{\'o}n, and
  Carlos Balaguer, `Past, present and future of robotic tunnel inspection',
  {\em Automation in Construction}, {\bf 59},  99--112, (2015).

\bibitem{a_rose2014supervised}
Paden Rose, Bryant Aaron, Dan~E Tamir, Lucy Lu, Jiong Hu, and Hongchi Shi,
  `Supervised computer-vision-based sensing of concrete bridges for
  crack-detection and assessment', Technical report, (2014).

\bibitem{simonyan2014very}
Karen Simonyan and Andrew Zisserman, `Very deep convolutional networks for
  large-scale image recognition', {\em arXiv preprint arXiv:1409.1556}, (2014).

\bibitem{sinha2006morphological}
Sunil~K Sinha and Paul~W Fieguth, `Morphological segmentation and
  classification of underground pipe images', {\em Machine Vision and
  Applications}, {\bf 17}(1), ~21, (2006).

\bibitem{sterling2010state}
R~Sterling, J~Simicevic, E~Allouche, W~Condit, and L~Wang, `State of technology
  for rehabilitation of wastewater collection systems', {\em Rep.
  EPA/600/R-10/078, US Environmental Protection Agency, Washington, DC.(Mar.
  25, 2012)}, (2010).

\bibitem{yamaguchi2008image}
Tomoyuki Yamaguchi, Shingo Nakamura, Ryo Saegusa, and Shuji Hashimoto,
  `Image-based crack detection for real concrete surfaces', {\em IEEJ
  Transactions on Electrical and Electronic Engineering}, {\bf 3}(1),
  128--135, (2008).

\bibitem{yosinski2014transferable}
Jason Yosinski, Jeff Clune, Yoshua Bengio, and Hod Lipson, `How transferable
  are features in deep neural networks?', in {\em Advances in neural
  information processing systems}, pp. 3320--3328, (2014).

\bibitem{zakeri2017image}
H~Zakeri, Fereidoon~Moghadas Nejad, and Ahmad Fahimifar, `Image based
  techniques for crack detection, classification and quantification in asphalt
  pavement: a review', {\em Archives of Computational Methods in Engineering},
  {\bf 24}(4),  935--977, (2017).

\end{thebibliography}
\end{document}